\documentclass[10pt,twocolumn,letterpaper]{article}

\usepackage{iccv}              % To produce the CAMERA-READY version
\definecolor{iccvblue}{rgb}{0.21,0.49,0.74}
\usepackage[pagebackref,breaklinks,colorlinks,allcolors=iccvblue]{hyperref}
\usepackage{graphicx}
\usepackage{caption}
\usepackage{multirow}
\usepackage{makecell}
\usepackage{colortbl}
\usepackage{tabularx}
\usepackage{fontawesome}

\def\paperID{13621} % *** Enter the Paper ID here
\def\confName{ICCV}
\def\confYear{2025}

\title{Griffon v2: Advancing Multimodal Perception with High-Resolution Scaling and Visual-Language Co-Referring}

\author{Yufei Zhan\textsuperscript{1, 2}, Shurong Zheng\textsuperscript{1, 4}, Yousong Zhu\textsuperscript{1, \faEnvelopeO}, Hongyin Zhao\textsuperscript{1}, Fan Yang\textsuperscript{1, 4},\\ Ming Tang\textsuperscript{1,  2}, Jinqiao Wang\textsuperscript{1, 2, 3, 4, \faEnvelopeO}\\
{\textsuperscript{1} Foundation Model Research Center, Institute of Automation,}\\
{Chinese Academy of Sciences, Beijing, China}\\
{\textsuperscript{2} School of Artificial Intelligence, University of Chinese Academy of Sciences, Beijing, China}\\
{$^3$ Peng Cheng Laboratory, Shenzhen, China}\;\;\;
{$^4$ Wuhan AI Research, Wuhan, China}\\
{\tt\small \{zhanyufei2021, zhengshurong2023, zhaohongyin2020, yangfan\_2022\}@ia.ac.cn}\\
{\tt\small \{yousong.zhu, tangm, jqwang\}@nlpr.ia.ac.cn}\\
}

\begin{document}
\maketitle
\begin{abstract}
Large Vision Language Models have achieved fine-grained object perception, but the limitation of image resolution remains a significant obstacle to surpassing the performance of task-specific experts in complex and dense scenarios. Such limitation further restricts the model's potential to achieve nuanced visual and language referring in domains such as GUI Agents, counting, \textit{etc}. To address this issue, we introduce a unified high-resolution generalist model, Griffon v2, enabling flexible object referring with visual and textual prompts. To efficiently scale up image resolution, we design a simple and lightweight down-sampling projector to overcome the input tokens constraint in Large Language Models. This design inherently preserves the complete contexts and fine details and significantly improves multimodal perception ability, especially for small objects. Building upon this, we further equip the model with visual-language co-referring capabilities through a plug-and-play visual tokenizer. It enables user-friendly interaction with flexible target images, free-form texts, and even coordinates. Experiments demonstrate that Griffon v2 can localize objects of interest with visual and textual referring, achieve state-of-the-art performance on REC and phrase grounding, and outperform expert models in object detection, object counting, and REG. Data and codes are released at \url{https://github.com/jefferyZhan/Griffon}.
\end{abstract}
\section{Introduction}
\label{sec:intro}
Large Vision Language Models (LVLMs) \cite{liu2023llava, alayrac2022flamingo, zhu2023minigpt, instructblip} show promising performance in region-level tasks like Referring Expression Comprehension (REC) after the breakthrough in image-text understanding \cite{antol2015vqa, chen2015microsoft} and reasoning \cite{johnson2017clevr}. In particular, models like Griffon \cite{zhan2023griffon} have demonstrated more compelling perception capability in object detection tasks. This has further spurred the development of flexible references of objects beyond only textual descriptions for better user interaction.
\begin{figure*}
    \centering
    \includegraphics[width=0.85\linewidth]{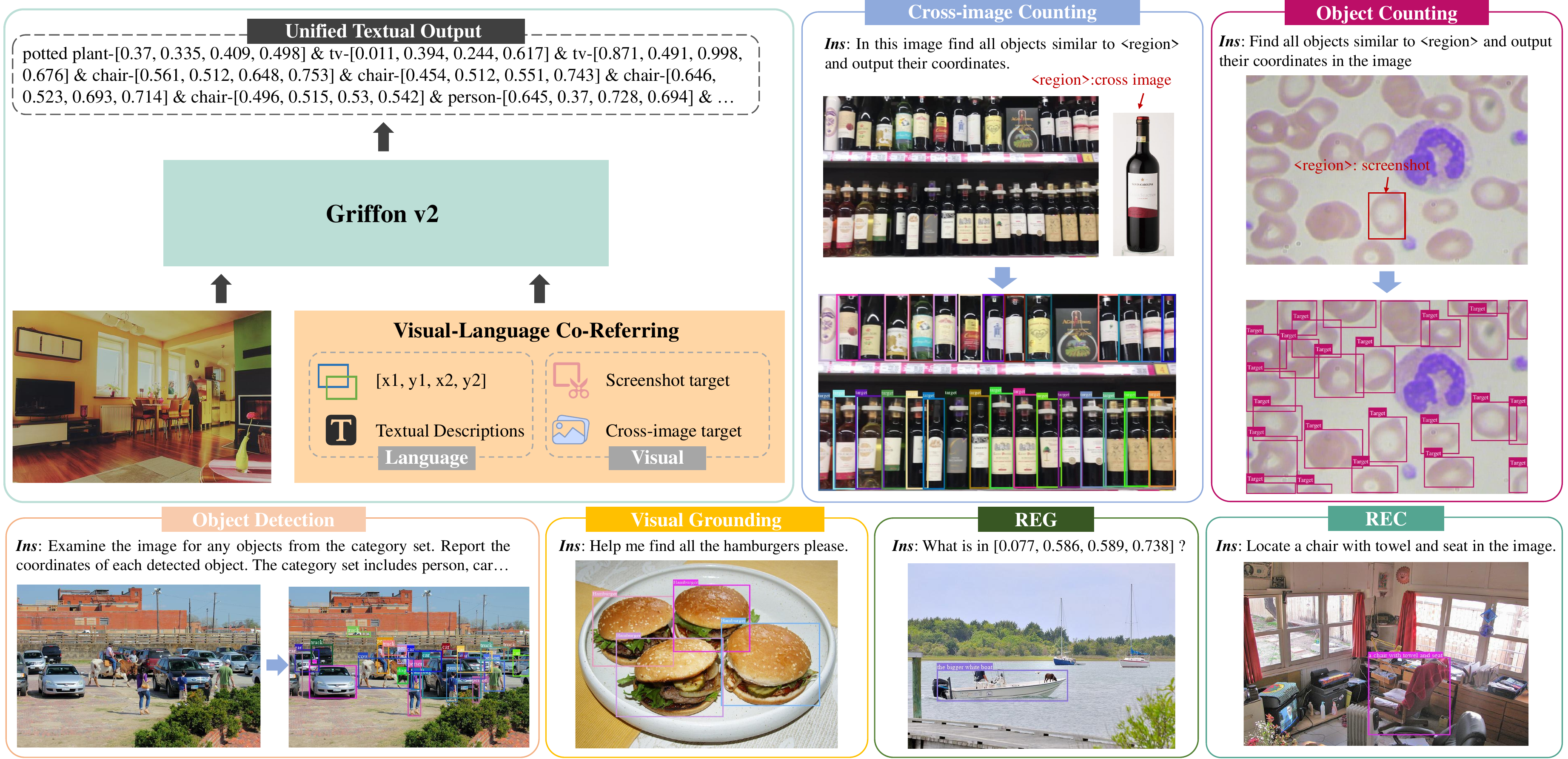}
    \caption{Overview of Griffon v2. Griffon v2 enables high-resolution input and seamless visual-language co-referencing for LVLMs, allowing users to refer to objects through coordinates, textual descriptions, screenshots, or cross-image modes. It excels in localizing arbitrary objects and generating descriptions with precise co-referencing across diverse scenarios.}
    \label{fig: benchmark}
\end{figure*}

Despite these progresses, current LVLMs still meet the bottleneck in surpassing the task-specific experts in fine-grained tasks, tripped by the image resolution. In other words, they can hardly capture the nuanced visual details, leading to a plethora of fact-conflicting hallucinations. It is particularly evident when dealing with low-resolution scenarios such as answering region-based questions without basis \cite{qi2024cogcom}, failing in small words in characters-related tasks \cite{liu2023hidden}, or providing incorrect counting results \cite{li2023otterhd}.

To address this issue, recent works have explored resolution enhancement and flexible visual-language referring in LVLMs. On the one hand, previous methods \cite{chen2023pali, chen2023palix} adopt a progressive training approach to gradually improve the resolution. However, the maximum input length of Large Language Models (LLMs) imposes a constraint on achieving higher image resolutions. Additionally, some approaches \cite{lin2023sphinx, li2023monkey, liu2024llavanext} divide the image into smaller patches and encode them separately for zooming in. This division-based design proves sub-optimal multimodal perception ability, which loses the contexts and edge details of patches and increases computation complexity \cite{li2023monkey}. On the other hand, prior research has predominantly studied various forms of references \cite{chen2023shikra, zhang2023gpt4roi, cai2023making} to improve the understanding of specific image content based on low-resolution images (\textit{e.g.} 224 or 448). However, these methods often excel at perceiving prominent, and image-level objects, but fall short in accurately localizing and describing fine-grained local objects. Additionally, singular visual or language prompts alone either lack conversational abilities or are constrained by linguistic descriptions \cite{jiang2023t}, failing to provide a comprehensive user-friendly interactive experience.

In this paper, we propose Griffon v2 to unveil the direct high-resolution design and endow it with locating any objects of interest with visual-language co-referring. Instead of partitioning the high-resolution image into smaller patches, we employ a high-resolution visual encoder to directly extract representations, and design a simple and lightweight down-sampling projector with strided convolution to compress the length of visual tokens. The compressed visual tokens are then trained to align with text tokens and fed into LLM for further fusion like LLaVA \cite{liu2023llava}. Compared to complex resampler structure\cite{lin2023sphinx} and image partitioning methods mode\cite{li2023monkey, liu2024llavanext}, the proposed direct high-resolution pipeline preserves context and advances dense instance-level tasks a lot with competitive VQA performances. It is also parameter-efficient and computationally concise. More importantly, we build a visual-language co-referring paradigm to enhance the model's fine-grained perception of high-resolution inputs, greatly expanding the model's applicability. It supports local cropped images, texts, and coordinates prompting, and outputs coordinates of target objects or corresponding text descriptions, providing various interactive abilities, thereby mitigating the conversational deficiencies of singular visual prompting and the potential expressive ambiguity of textual prompting. Finally, we collected 12M publicly available localization data for pre-training and 900K instruction data for fine-tuning. We achieve advanced results in the REC task, phrase grounding task, and Referring Expression Generation (REG) task. Notably, our model outperforms several object detection and object counting expert models for the first time.

In summary, our main contributions are:
\begin{enumerate}
    \item We propose a high-resolution multimodal perception model for better local understanding and less global context loss, better suited for challenging tasks such as intensive multi-object detection and counting in complex scenarios.
    \item We introduce a visual-language co-referring structure, which broadens the model's scope of application and provides various interaction modes.
    \item We have conducted experiments on a wide range of localization-related tasks and demonstrated state-of-the-art performance on REC, phrase grounding, and REG. We surpass expert models in object detection and object counting task quantitatively and qualitatively.
\end{enumerate}

\section{Related Work}
\label{sec:related}
\begin{figure*}[t]
  \centering
  \includegraphics[width=0.85\linewidth]{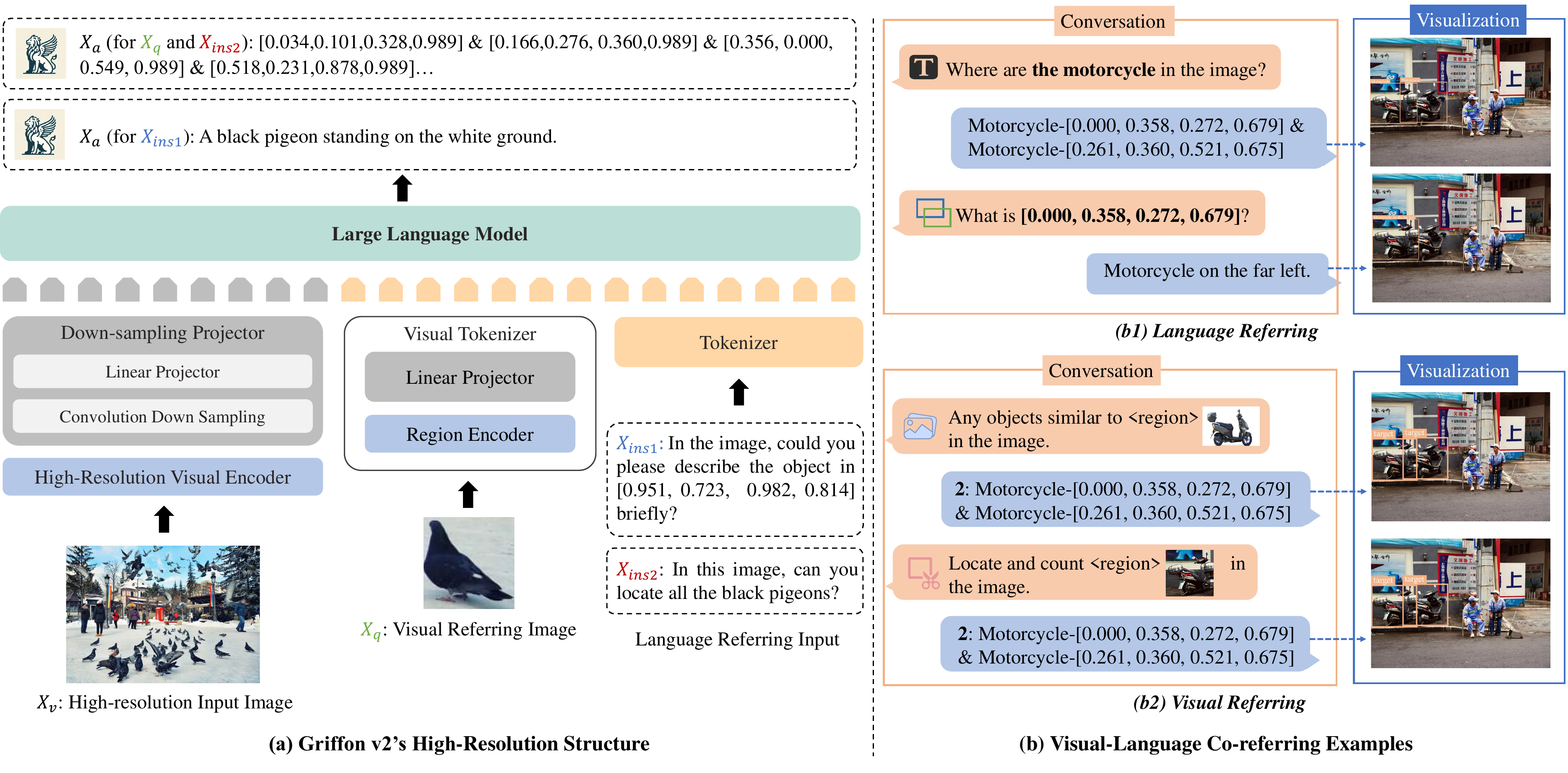}
   \caption{Structure of the proposed high-resolution Griffon v2 with visual-language co-referring.}
   \label{fig: structure}
\end{figure*} 

\subsection{LVLMs with Localization} With the significant advancements made in LVLMs\cite{liu2023llava,alayrac2022flamingo,instructblip, achiam2023gpt}, object localization as an essential foundation visual task has been explored with LVLMs. One branch of these methods \cite{yang2023auto, shen2023hugginggpt, zhao2023bubogpt, liang2023taskmatrix} focuses on invoking the detection expert models through an API. When the LVLM requires detection, the LLM issues instructions to call the detection model and receives the results for further processing. This approach increases the complexity of the model and can not enjoy the unified structure with LVLMs. Another set of methods \cite{chen2023shikra, bai2023qwen, you2023ferret, chen2023pali} focuses on fine-grained single-object localization, \textit{i.e.} REC. These methods transform REC task data into instruction data and encode the coordinates with different representation approaches, enabling the LVLM itself to possess localization capabilities. However, methods like Ferret \cite{you2023ferret} fall short in  more complex tasks like object detection and more challenging scenarios with lots of small objects under the low-resolution. Until the appearance of Griffon\cite{zhan2023griffon}, which supports coarser granularity object detection, the object localization capacity of LVLM is extended to multi-object and arbitrary granularity. Our model further proposes a high-resolution structure that elevates the detection and object counting capability of LVLM beyond expert models, which has also proved more effective and efficient than existing division-based resolution-enhanced methods \cite{li2023monkey, lin2023sphinx}. By enhancing the object detection and enabling object counting, our work address a broader spectrum of granularity in comprehension, perception and localization.

\subsection{Object Referring in Multimodal Models}
Since the application of LLMs \cite{touvron2023llama, achiam2023gpt} in large multimodal models, textual descriptions have become the most straightforward method for object referring. Users naturally utilize this method to ask questions without even paying special attention, such as ``What is in front of this cat?'' Despite its simplicity, it struggles to distinctly refer to specific classes of objects that are difficult to describe in dense and complex scenes, such as the cell shown in Figure \ref{fig: benchmark}. The growing importance of region comprehension has sparked the exploration of spatial references in complex and dense scenes. Current works have proposed spatial referrals with textual coordinates \cite{chen2023shikra, bai2023qwen, chen2023minigptv2}, learnable embeddings \cite{peng2023kosmos}, and Region of Interest features \cite{zhang2023gpt4roi}. Some approaches \cite{cai2023making, you2023ferret, achiam2023gpt} upgrade from the user-friendly perspective, supporting arrows, marks, and so on.  Although these methods enhance the convenience of user interaction, they primarily focus on designing visual prompts to distinguish one specific object from others and are mostly applied in text-output VQA tasks. In contrast, our model uses visual referring to represent a class of objects, not limited to distinction of individual objects. Additionally, we introduce bounding box output to enable dense scenario understanding in LMMs for the first time. The proposed visual-language co-referring architecture empowers our model to handle various tasks while remaining user-friendly.
\section{Methodology}
\label{sec:formatting}
In this section, we start with a brief overview of our Griffon v2. As the key foundation, we first describe the high-resolution structure design. Then, we represent the visual-language co-referring and the training pipeline.
\subsection{Overview}
As depicted in Figure \ref{fig: structure}(a), Griffon v2 employs an auto-regression paradigm, seamlessly integrating referring and grounding tasks within a unified language modeling objective. For a given input image $X_v$, we leverage a visual encoder, adapted to high-resolution by bilinear interpolation from the pre-trained EVA2-CLIP-L/14@336 \cite{sun2023eva} model to extract high-resolution visual features (\textit{e.g.} over 1K). Simultaneously, the designed down-sampling projector transforms these image features into visual embedding tokens $H_v$. In cases where a user customizes a visual prompt $X_{q}$ with a screenshot or an additional target image, region embeddings $H_{q}$ are derived from the visual referring image using the visual tokenizer. Meanwhile, we inherently support language references, $X_{ins}$, to identify objects, such as textual descriptions and coordinates, which are projected into text embeddings as $H_{ins}$ after tokenization. Subsequently, the visual prompt or text embeddings together with image embeddings undergo processing using the LLM, specifically Llama2-13B \cite{touvron2023llama}, resulting in the generation of the desired answer $X_a$. Notably, the representation of bounding boxes eschews the use of special tokens and specialized heads and instead incorporates textual numerical values, thereby maintaining simplicity.

\subsection{Efficient High-Resolution Scaling Design}
\label{sub: high-res}
High resolution has been empirically proven to enhance the accurate interpretation of nuanced visual features, thereby benefiting both visual tasks and vision-language tasks. In previous methods \cite{li2023monkey}, images are commonly partitioned into smaller patches, enlarging their total resolution to 762 or 896. However, this approach faces limitations in terms of contextual coherence and the representation of edge details among patches. While progressive scaling methods can extend resolution to 784 (patch size$=$14) within the constraint of 4096 input tokens, they are unsuitable for handling outputs involving extensive long texts and significantly increase the computation complexity. 

To address these challenges, we introduce a high-resolution structure, incorporating a down-sampling projector capable of supporting a resolution of more than 1K. Upon receiving an input image of arbitrary dimensions, we resize it to a predefined input height and width, denoted as $X_v \in {R}^{H \times W \times 3}$. Instead of employing image division, we encode the image using a trainable high-resolution visual encoder adapted from a pre-trained model by bilinear interpolation, yielding visual features $Z_v = f_V(X_v)$. In contrast to low-resolution fixed encoders, our approach excels in capturing finer details. As illustrated in the left part of Figure \ref{fig: structure}(a), we subsequently introduce a lightweight down-sampling projector to abstract essential features with compression under the guidance of the training objective and connect visual features to word embedding space. Specifically, a strided convolution layer ($Conv$) is simply applied to down-sample the features by $S$, and a projection matrix $W$ is utilized to convert $Z_v$ into visual embedding tokens:
\begin{equation}
    H_v = W \cdot Conv(Z_v),\;with\;Z_v = f_V(X_v).
\end{equation}

\begin{figure}[t]
  \centering
  \includegraphics[width=0.9\linewidth]{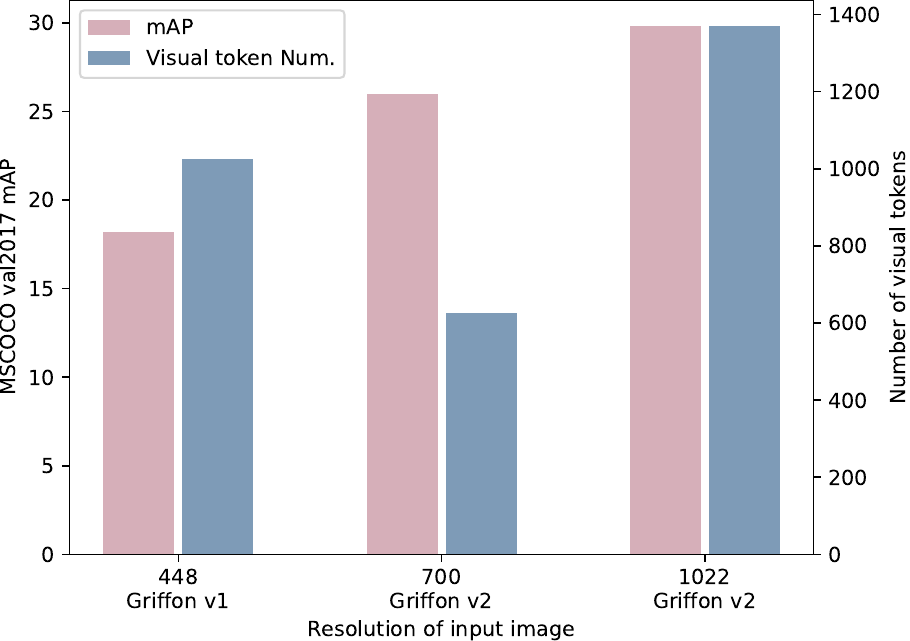}
   \caption{Performance and visual token number comparison of different resolutions under the same training data. The 448-resolution model, \textit{i.e.} Griffon v1, employs a two-layer linear projector, whereas the others utilize our down-sampling projector (Stride$=$2, Kernel Size$=$3). Our method is capable of extracting features while preserving details, thereby reducing token redundancy.} 
   \label{fig: resolution}
\end{figure}

It can be inferred that the final number of visual tokens is quadratically related to the input size and the convolution stride, \textit{i.e.} $H, W$, and $S$. To demonstrate the effectiveness of our design, we employ the convolution layer with kernel size at 3 and stride at 2 and investigate several different resolutions in Figure \ref{fig: resolution}. As depicted in Figure \ref{fig: resolution}, comparing the results at resolutions of 700 and 448 which use two-layer linear projection \cite{liu2023llava} without reduction, using the resolution of 700 with our structure shows a higher mAP and fewer tokens. This indicates that our method is capable of extracting features while preserving details, thereby reducing token redundancy. It is also observed that further increasing the resolution can lead to improved performance.

\subsection{Visual-Language Co-Referring}
\label{sub: VL-refer}
Recognizing the limitations of singular visual or language referring demonstrated before, lack of conversational ability, and potential ambiguity respectively, we capitalize on CLIP's adeptness in extracting robust semantic features to facilitate visual-language co-referring. This approach mitigates these drawbacks and finds broader applicability under high-resolution conditions.

\textbf{Visual Referring.} While various existing methods propose diverse visual reference forms, ranging from marking on the raw image to object masking, either of them alters the raw image content \cite{achiam2023gpt} or requires a specially designed extractor \cite{you2023ferret}. To strike a balance between efficiency and referring accuracy, we opt for a straightforward method, using visual referring images to represent the target of interest. Specifically, as depicted in the middle of Figure\ref{fig: structure}(a), given a user-provided referring image $X_{q}$, we employ a region encoder, \textit{i.e.} EVA-CLIP-L/14@224 \cite{sun2023eva} pre-trained ViT to extract visual prompt features $Z_{q}$. To align the LLM's understanding of visual references with textual references, we project the region features into the word embedding space, yielding the visual prompt token $H_{q}$. Based on the robust semantic representation capability of $[CLS]$ token, pre-trained on a dataset of billions of instances \cite{radford2021learning}, we discard other tokens following previous applications \cite{radford2021learning, sun2023eva} and utilize the $[CLS]$ token to represent the specific object. This approach allows us to refer to the object at the category level, eliminating overly specific information from pixel patches and maintaining the stability of reference. To accommodate diverse conversation scenarios, we use ``$<$region$>$'' in instructions to represent the referred object as shown in Figure \ref{fig: structure}(b). Before inputting into the LLM, we replace this placeholder with the embedding of the referred object, seamlessly integrating it for optimal coherence.
% \begin{figure*}[t]
%   \centering
%   \includegraphics[width=0.95\linewidth]{fig/VL-referring_v2.pdf}
%    \caption{{Examples of visual-language co-referring usage in conversations.}} 
%    \label{fig: example}
% \end{figure*} 

\textbf{Language Referring.} Textual referring is an inherent capability of LLM-based multimodal models. Despite the simplicity of textual referring, it faces the challenge of resulting in wrong answers with ambiguous referring. This has naturally prompted us to consider whether associating texts with regions could allow LLMs to comprehend the corresponding object based on textual-region reference and thereby fulfill task instructions and keep the simplicity advantage. Therefore, we further support textual coordinate references as illustrated in Figure \ref{fig: structure}(b). In addition to the detailed description, an object can also be referred to by its top-left and bottom-right coordinates. In this case, it is processed the same as textual descriptions, involving tokenization and embedding into the word space as $H_{ins}$, depicted at the right side of Figure \ref{fig: structure}(a). Through training with instruction-following data whose object references are transformed into textual coordinates, we demonstrate that LLMs can accurately refer to objects in complex and dense scenes based on the textual coordinates of objects. Through collaborative referring involving both text and vision, our approach achieves optimal referring performance and flexible interaction.

\subsection{Training Pipeline}
\label{sub: train}
To understand different users' intents and finish diverse tasks with high-quality results, we adopt a three-stage end-to-end training procedure for Griffon v2 as follows guided by \cite{liu2023llava, zhan2023griffon}.
\begin{table}[h]
    \centering
    \fontsize{9pt}{\baselineskip}\selectfont
    \setlength{\tabcolsep}{2pt}
    \begin{tabular}{cll}
        \toprule
         Stage &  Annotation Type & Sources\\
         \midrule
        \MakeUppercase{\romannumeral1} &  Image-text & LLaVA \\
        \midrule
         \multirow{5}{*}{\MakeUppercase{\romannumeral2}} & Object Detection& Objects 365, MSCOCO \\
         & REC/REG& Visual Genome, RefCOCO series \\
         & Visual Grounding& V3Det, LVIS, Flickr30K Entities \\
         & Object Counting& CA-44, OpenImages, Self-collected \\
         & Non-existing Judging & LVIS\\
         \midrule
         \MakeUppercase{\romannumeral3} &  Instruction-following & Build from stage 2 \\
         \bottomrule
    \end{tabular}
    \caption{Data statistics of different stages. We convert the data of different annotations to Prompt/Instruction-Answer form. We will release the self-collected and processed data in Stage \MakeUppercase{\romannumeral2} and \MakeUppercase{\romannumeral3}. More details are listed in the supplements.}
    \label{data:stat}
\end{table}

\textbf{Stage \MakeUppercase{\romannumeral1:}} {\bf High-resolution Vision-Language Alignment.} Feature alignment before pretraining has been widely utilized to achieve better training efficiency. We adopt this strategy to connect the high-resolution visual encoder with the LLM using 558K image-text pairs\cite{liu2023llava}. The high-resolution image encoder and LLM parameters remain frozen, with only the down-sampling projector trainable. Without visual prompts in these image-text pairs, the visual prompt tokenizer is not trained in Stage \MakeUppercase{\romannumeral1}.

\textbf{Stage \MakeUppercase{\romannumeral2:}} {\bf Co-referring Multi-Tasks Pre-training.} 
Building on the basic understanding of visual content achieved in Stage \MakeUppercase{\romannumeral1}, we further pre-train the entire model with a diverse dataset comprising 12M multi-task instances involving both visual and textual referring. This stage aims to enhance fine-grained perception and localization capabilities, and visual-language referring abilities. As detailed in the data composition in Table \ref{data:stat}, our textual-referring data is curated from various task datasets covering over 70K object categories, such as Visual Genome \cite{krishna2017visual} and V3Det\cite{wang2023v3det}. Additionally, we build a dataset with public and self-collected object counting data from 10 different domains, spanning aerial photography, agriculture, industry, and more, to create our visual-referring data. The diverse textual categories and visual domains contribute to the model's generalization capability. During this stage, we train the entire network while keeping the region encoder in the visual prompts tokenizer frozen.

\textbf{Stage \MakeUppercase{\romannumeral3:}} {\bf Intent-Enhanced Instruction Tuning.} Following the pretraining of stage \MakeUppercase{\romannumeral2}, the model gains the ability to locate and describe objects of interest with free-form texts and flexibly obtained visual referring images. It can be customized and enhanced by users to achieve specific tasks as a foundation model. To further enhance its understanding ability of user's intents, we finetune the model with nearly 900K instruction-following data built from stage \MakeUppercase{\romannumeral2} and more diverse instruction prompts for different tasks detailed in supplements. Moreover, to preserve the refined visual feature extraction and prevent forgetting, we keep the high-resolution visual encoder and region encoder frozen and train the LLM together with the projectors.
\section{Experiments}
\label{sec: exp}
\begin{table*}[t]
    \centering
    \fontsize{9pt}{\baselineskip}\selectfont
    \begin{tabular}{llccccccccc}
        \toprule
        {Type}& {Model}& Res. & Epochs &$mAP$&$AP_{50}$&$AP_{75}$&$AP_S$&$AP_M$&$AP_L$\\
         \midrule
         \multirow{5}{*}{Specialists}
         &Faster RCNN-FPN\cite{ren2015faster} & 1022 & 12 & 37.9 & 58.6 & 40.9 & 20.4 & 41.1 & 50.3 \\
         &Faster RCNN-C4\cite{ren2015faster}& 1022 & 12 & 35.6 & 55.7 & 37.8 & 17.0 & 40.6 & 50.3 \\
         & DAB-DETR\cite{liu2022dab} & 1333 & 12 & 38.0 & 60.3 & 39.8 & 19.2 & 40.9 & 55.4  \\
         & Pix2Seq\cite{chen2021pix2seq} & 1333 & 300 & 43.0 & 61.0 & 45.6 & 25.1 & 46.9 & 59.4 \\
         & DETR\cite{carion2020end} & 1333 & 500 & 42.0 & 62.4 & 44.2 & 20.5 & 45.8 & 61.1 \\
         \midrule
         & Griffon-13B\cite{zhan2023griffon} & 448 & 1 & 24.8 & 40.6 & 25.1 & 5.9 & 25.5 &  48.7 \\
         & Lumen\cite{jiao2025lumen} & 448 & - & 35.3 & 53.2 & 35.8 & - & - & - \\
         & Qwen2.5-VL-7B\cite{bai2025qwen25vl} & dynamic & - & 16.2 & 25.0 & 16.7 & 5.9 & 16.8 & 33.8 \\
         & InternVL2.5-8B\cite{chen2025internvl25} & dynamic & - & 11.9 & 19.4 & 12.1 & 1.9 & 11.2 & 24.2 \\
         % & PaliGemma2-10B & 896 &  &  &  &  &  &  &  \\
        \multirow{-6}{*}{Generalist} & Griffon v2 & 1022 & 1 & \textbf{38.5} & \textbf{54.3} & \textbf{41.2} & \textbf{19.4} & \textbf{43.2} & \textbf{57.6}  \\
         % & \textbf{Ours*} & 1022 & 1 & 40.5 & 57.9 &  43.0 & 22.0  &  45.5 &  60.9  \\
         \bottomrule
    \end{tabular}
    \caption{Object detection results on MSCOCO val2017 \cite{lin2014microsoft}. Griffon v2 shows a significant improvement over advanced LVLMs.}
    \label{tab: detection results}
\end{table*}

\begin{table}[t]
    \centering
    \fontsize{9pt}{\baselineskip}\selectfont
    \begin{tabular}{llcc}
        \toprule
        Type& Model& MAE($\downarrow$) & NAE($\downarrow$)\\
         \midrule
         \multirow{3}{*}{Specialists} & FamNet\cite{m_Ranjan-etal-CVPR21}  & 68.5 & 2.3\\
         & FSDetView\cite{xiao2022few}  & 29.0 & 0.8 \\
         & Counting-DETR\cite{nguyen2022few} & 23.5 & 0.6 \\
        \midrule
        \multirow{3}{*}{Generalist} 
        & Qwen2.5-VL-7B\cite{bai2025qwen25vl}  & 33.0 & 0.9 \\
        & InternVL2.5-8B\cite{chen2025internvl25}  & 28.5 & 0.9 \\
        & Griffon v2  & \textbf{20.3} & \textbf{0.5} \\
         % & \textbf{Ours*} & 1022 & 1 & 40.5 & 57.9 &  43.0 & 22.0  &  45.5 &  60.9  \\
         \bottomrule
    \end{tabular}
    \caption{Object counting results on the dense FSCD-LVIS unseen test set \cite{nguyen2022few}. MAE stands for Mean Average Error, and NAE for Normalized Relative Error.}
    \label{tab:counting results}
\end{table}

\subsection{Implementation Details}
Following the widely used configuration of ResNet\cite{he2016deep}, we set the convolution layer of the down-sampling projector with a kernel size of 3, a stride of 2, and a padding of 1. For the chosen image resolution, we consider both the patch size of pre-trained CLIP and the token length of our data under the constraint of Llama2, supporting a maximum of 4096 input tokens. We follow the previous methods to utilize the L-size CLIP model, whose patch size is usually 14 or 16. As the average textual token length of counting data is approximately 2500, the maximum achievable resolution is 1022 (patch size = 14) or comparable 1024 (patch size = 16), corresponding to 1369 tokens or 1024 tokens respectively. As shown in Table \ref{aba: pretrain}, as the EVA-CLIP-L/14 has the best performance, we set the resolution to 1022. We initialize the visual encoder with adapted EVA2-CLIP-ViT-L/14@336 by position embedding interpolation and LLM with Llama2, leaving the down-sampling projector and projector of visual tokenizer randomly initialized. More details are demonstrated in the supplements.

\subsection{Complex Detection and Counting}
Object detection and counting are essential visual perception tasks, presenting significant challenges with multiple categories and dense objects. We evaluate models on these two tasks as the first LVLM and demonstrate our fine-grained perception ability in complex and dense scenarios.

\textbf{Object Detection.} The object detection task is evaluated on MSCOCO val2017 \cite{lin2014microsoft} using textual references and Griffon-13B's prompt. We input all test categories simultaneously for each image and calculate the confidence score for each prediction following \cite{zhan2023griffon}. As illustrated in Table \ref{tab: detection results}, we outperform existing expert models, including Faster RCNN (C4 and FPN)\cite{ren2015faster} and DAB-DETR\cite{liu2022dab}, with fewer training epochs and lower input resolution. Moreover, we outperform the first pure LVLM generalist Griffon-13B under the same data and training settings as depicted in Figure \ref{fig: resolution}, and also achieve substantial improvements across all metrics compared to the generalist Griffon-13B overall. Compared to Lumen\cite{jiao2025lumen}, which uses a task-specific decoder to output detection results, our Griffon v2 surpasses by 3.2\% on mAP score, showcasing the superior detection capability of ours. The latest Qwen2.5-VL\cite{bai2025qwen25vl}, in contrast, faces challenges in object detection, as the partition-based high-resolution structure may lose edge details and contexts of patches and is not the optimal solution for intensive multi-object localization under high-resolution.

\begin{table*}[t]
    \centering
    \setlength{\tabcolsep}{3.5pt}
    \fontsize{9pt}{\baselineskip}\selectfont
    \begin{tabular}{clccccccccccccccc}
        \toprule
         \multirow{2}{*}{Type} & \multirow{2}{*}{{Model}} & \multicolumn{14}{c}{ODINW}&\multirow{2}{*}{RefCOCO/+/g} \\
         \cmidrule(r){3-16}
          &  & \tiny \rotatebox{75}{AerialDrone} & \tiny \rotatebox{75}{Aquarium} & \tiny \rotatebox{75}{Rabbits} & \tiny \rotatebox{75}{EgoHands} & \tiny \rotatebox{75}{Mushrooms} & \tiny \rotatebox{75}{Packages} & \tiny \rotatebox{75}{PascalVOC} & \tiny \rotatebox{75}{pistols} & \tiny \rotatebox{75}{pothole} & \tiny \rotatebox{75}{Raccoon} & \tiny \rotatebox{75}{Shellfish} & \tiny \rotatebox{75}{thermal} & \tiny \rotatebox{75}{Vehicles} & \tiny \rotatebox{75}{AVG} & AVG\\
         \midrule
         \multirow{2}{*}{\rotatebox{90}{Spec.}}
         &MDETR\cite{kamath2021mdetr}  & 0.6 & 1.7 & 66.5 & 5.9 & 39.8 & 63.6 & 5.6 & 15.9 & 12.7 & 50.6 & 8.1 & 4.5 & 13.4 & 22.2 & 83.4 \\

         &G-DINO-L\cite{liu2023grounding} & 12.6 & 28.1 & 71.7 & 52.0 & 72.3 & 63.9 & 66.0 & 71.4 & 30.4 & 65.8 & 62.5 & 21.3 & 62.7 & 52.4 & 86.6\\

         \midrule
         \multirow{7}{*}{\rotatebox{90}{Gen.}}

         &Ferret-13B\dag\;\cite{you2023ferret} & 0 & 4.3 & 59.8 & 1.5 & 6.1 & 40.1 & 35.2 & 41.5 & 3.9 & 49.5 & 29.5 & 36.5 & 44.4 & 27.1 & 85.6\\
         &Griffon-13B\cite{zhan2023griffon} & - & - & - & - & - & - & - & - & - & - & - & - & - & - & 84.0\\
         &InternVL2.5-8B\dag\;\cite{chen2025internvl25} & 0 & 6.9 & 38.5 & 0.2 & 26.7 & 16.4 & 37.0 & 29.2 & 1.1 & 46.6 & 28.5 & 3.8 & 27.1 & 20.2 & 87.6\\
         &Qwen2.5-VL-3B\cite{bai2025qwen25vl} & 6.2 & 16.4 & 75.0 & 24.6 & 8.3 & \textbf{66.6} & 52.0 & 42.3 & \textbf{10.2} & \textbf{47.7} & 36.7 & 40.7 & 57.1 & 37.2 & 85.0\\
         &Qwen2.5-VL-7B\dag\;\cite{bai2025qwen25vl} & \textbf{7.8} & \textbf{20.3} & 73.5 & \textbf{32.2} & 7.0 & 57.6 & 49.8 & \textbf{48.5} & 7.4 & 40.1 & 42.7 & 38.0 & 56.3 & 37.0 & 86.6\\
         
         \cmidrule(r){2-17}
         &Griffon v2 & 5.4 & 18.2 & \textbf{75.1} & 25.7 & \textbf{63.7} & 62.1 & \textbf{62.0} & 43.1 & 6.4 & 39.8 & \textbf{44.1} & \textbf{43.1} & \textbf{57.6} & \textbf{42.0} & \textbf{90.0}\\

         \bottomrule
    \end{tabular}
    \caption{Visual Grounding and REC results. Spec. represents specialists, while Gen. represents generalists. RefCOCO/+/g utilizes the metric of $ACC @ 0.5$, while ODINW uses mAP metric. \dag\; indicates the results are reproduced to get results of all sets following the official Settings.}
    \label{tab:REC results}
\end{table*}

\textbf{Object Counting.} Object counting is conducted with visual references and tested on the unseen test classes set of FSCD-LVIS \cite{nguyen2022few} aiming for accurate counting and facilitating generalization comparisons. The visual reference is constructed by randomly selecting one example box from the set and screenshotting the region in the image. For Qwen2.5-VL-7B\cite{bai2025qwen25vl} and InternVL2.5-8B\cite{chen2025internvl25}, we query the model to count the number of the target following their trained settings. As depicted in Table \ref{tab:counting results}, we surpass existing classical expert models with lower MAE and NAE. Notably, our approach not only outputs the number but also provides the bounding boxes of detected objects. This marks the first time LVLMs achieve real expert-level detection and counting, showcasing the superiority of Griffon v2 and the generalization ability of our visual reference approach. By contrast, Qwen2.5-VL-7B and InternVL2.5-8B struggle to predict the accurate number of specific category of targets in dense scenarios, proving the effectiveness of our high-resolution scaling strategy and co-referring mechanism.

\subsection{Evaluation on Referring and Grounding}
\label{main}
Basic grounding and referring mainly include the visual grounding and REC tasks, typically involving one pre-existing object or a limited number of multiple targets. Griffon v2 is systematically compared with specialist and generalist models across these tasks.

As a fundamental task in visual grounding, the REC task has been extensively researched in the LVLMs as a basic localization task grounding a single object with textual description. We evaluate our model on the RefCOCO \cite{yu2016modeling}, RefCOCO+ \cite{yu2016modeling}, RefCOCOg \cite{nagaraja2016modeling} and it achieves an average score of 90.0. 
% RefCOCO/+/g is a relatively simple benchmark, as it primarily examines the fine-grained understanding and single-target localization capabilities. 
As illustrated in Table \ref{tab:REC results}, our model demonstrates excellent performance, outperforming InternVL2.5-8B\cite{chen2025internvl25} and Qwen2.5-VL-7B\cite{bai2025qwen25vl}.

To comprehensively assess the model's capabilities to generalize detection of uncommon categories in diverse real-world scenarios, we also conduct experiments on a more intensive object grounding benchmark, Object Detection in the Wild (ODinW)\cite{li2022glip}. ODinW contains more rare categories and multiple target locations in the wild.

As shown in Table \ref{tab:REC results}, our model outperforms recent advanced LVLMs across multiple sets especially when the query category corresponds to multiple targets, such as Mushrooms and PascalVOC. Compared to the latest Qwen2.5-VL-7B\cite{bai2025qwen25vl}, our model improved the average score by 5.0\%, narrowing the gap with specialist models. 

Beyond these two visual grounding tasks, we also evaluate phrase grounding and REG on Flickr30K Entities\cite{plummer2015flickr30k} and RefCOCOg \cite{nagaraja2016modeling} respectively. The detailed experimental setup and results are reported in Appendix \ref{sup:PG and REG}.

\subsection{Ablation Studies}
\label{ablation}
In the ablation studies below, by default, we mainly evaluate Griffon v2 on the object detection task on MSCOCO val2017 \cite{lin2014microsoft} and train the whole model with only the train2017 set after feature alignment.

\textbf{Different pre-trained visual encoders.} Existing methods utilize the visual encoder from the pre-trained CLIP model, and the performance of pre-trained models varies with different optimization methods. We compare EVA2-CLIP-L/14 \cite{sun2023eva}, original CLIP-L/14 \cite{radford2021learning} and SAM-CLIP-16 \cite{kirillov2023segment}. We apply bilinear interpolation to upscale the resolution from 336 to 1022, which closely matches the 1024 resolution of the pre-trained SAM-CLIP-16, due to the patch size difference (14 \textit{v.s.}16). As shown in Table \ref{aba: pretrain}, the EVA2-CLIP-based visual encoder outperforms the other two models with an average improvement of 2\%. Comparing the CLIP-based models using positional interpolation and the SAM-based model without interpolation, indicates that with our structure design, using positional interpolation to increase resolution can also work with even better performance.

\begin{figure*}[t]
  \centering
  \includegraphics[width=0.86\linewidth]{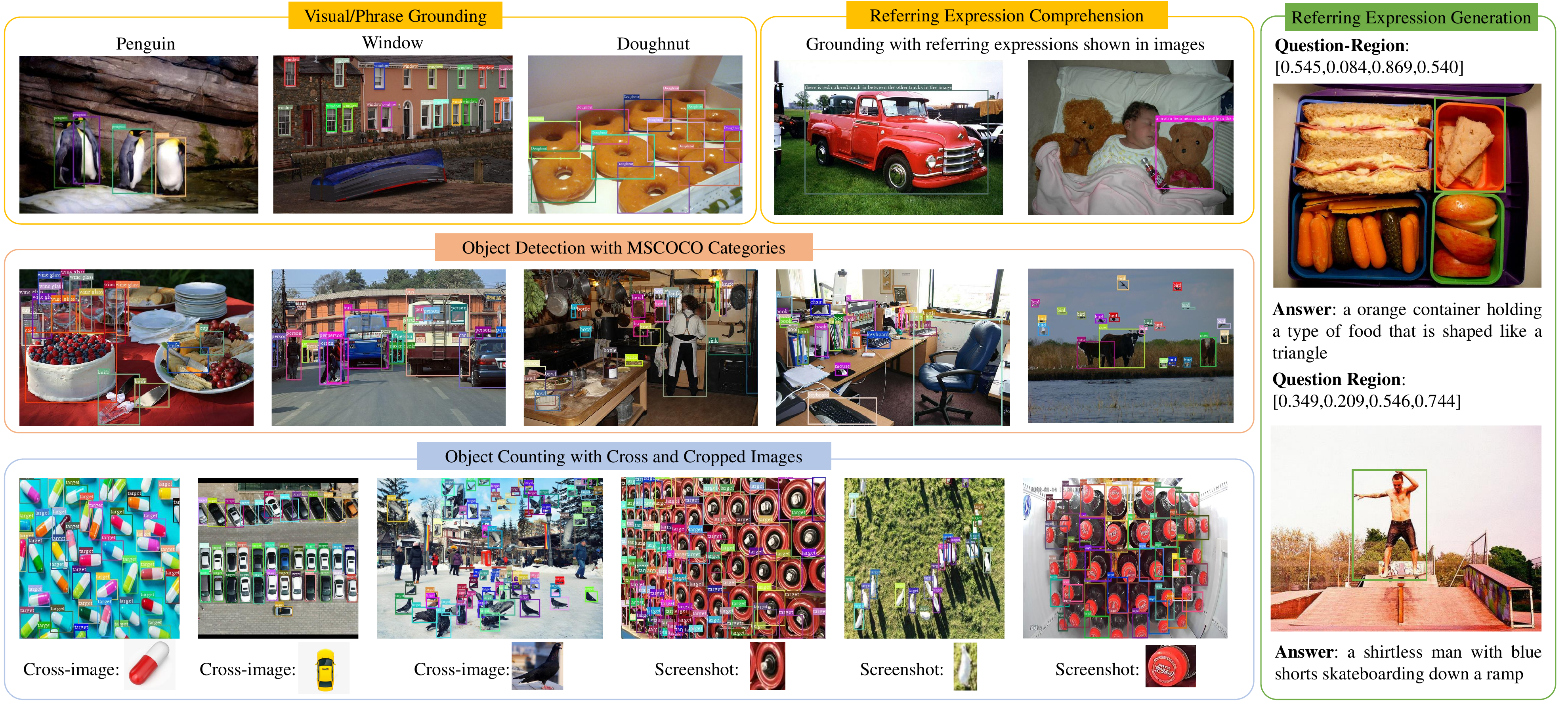}
   \caption{Visualization results of Griffon v2 across five vision and vision-language tasks.} 
   \label{fig: vis}
\end{figure*}

\textbf{High-resolution design.} Recent works\cite{lin2023sphinx, li2023monkey, liu2024llavanext} incorporate image partitioning to achieve dynamic high-resolution resolution. Though effective on multimodal benchmarks, it's unknown how it performs on fine-grained localization tasks. To further validate the effectiveness of our direct high-resolution scaling strategy, we use the same train data to conduct a fair comparison with the partition-based high-resolution approach. We use the multimodal data from LLaVA-NeXt\cite{liu2024llavanext} and compare with it, and use the COCO data to train Monkey\cite{li2023monkey} with similar solution for holistic comparison. As shown in Table \ref{aba: high-resolution}, Griffon v2 outperforms LLaVA-NeXt in general VQA and largely exceeding Monkey by 6.7\% in object detection. The results underscore the effectiveness of our high-resolution structure design, which can preserve the visual contexts and adapt to global understanding tasks.
% , 
\begin{table}[t]
    \centering
    % \footnotesize
    \fontsize{9pt}{\baselineskip}\selectfont
    \setlength{\tabcolsep}{2.8pt}
    \begin{tabular}{lc|cccccccc}
        \toprule
         Pretrained & $mAP$&$AP_{50}$&$AP_{75}$&$AP_S$&$AP_M$&$AP_L$ & $AR_{100}$\\
         \midrule
         CLIP  & 29.8 & 48.6 & 29.7 & 10.0 & 32.7 & \textbf{53.9} & 41.4 \\
         SAM  & 28.4 & 43.1 & 29.6 & 11.2 & 30.8 & 46.9 & 40.9 \\
         EVA2-CLIP & \textbf{31.9} & \textbf{50.4} & \textbf{32.7} & \textbf{14.3} & \textbf{38.6} & 52.3 & \textbf{43.8} \\
         \bottomrule
    \end{tabular}
    \caption{Ablations on different pre-trained visual encoders.}
    \label{aba: pretrain}
\end{table}

\begin{table}[t]
    \centering
    % \footnotesize
    \fontsize{9pt}{\baselineskip}\selectfont
    \setlength{\tabcolsep}{2.8pt}
    \begin{tabular}{c|ccccc}
        \toprule
         Model & VQAv2&ChartQA&MME$^{P}$&POPE&mAP\\
         \midrule
         LLaVA-NeXt\cite{liu2024llavanext}  & 80.9 & 62.2 & 1575 & 86.3 & - \\
         Monkey\cite{li2023monkey}  & - & - & - & - & 17.6 \\
         Griffon v2  & \textbf{81.9} & \textbf{69.8} & \textbf{1581} & \textbf{87.5} & \textbf{24.3} \\
         \bottomrule
    \end{tabular}
    \caption{Comparison with partition-based high-resolution structure on VQAs and object detection. We use the same train data with LLaVA-NeXt and Monkey respectively to ensure fairness.}
    \label{aba: high-resolution}
\end{table}
 
\textbf{Down-sampling structures.} Previous methods apply the resampler proposed in Flamingo \cite{alayrac2022flamingo} to down-sample the visual features with learnable tokens while increasing the resolution. We compare this approach with our designed down-sampling projector in terms of the performance and memory of this module during training. As depicted in Table \ref{aba: down}, our model achieves higher precision with less memory consumption, which is quite important for large-scale pre-training of LVLMs. While the resampler can extract semantic information for understanding tasks, it falls short in capturing fine-grained details for perception and localization tasks under the same training setting, necessitating more epochs \cite{carion2020end}.

\subsection{Qualitative Analysis}
\label{vis}
We further evaluate Griffon v2's performance across five tasks by presenting visualization results. As depicted in Figure \ref{fig: vis}, Griffon v2 consistently demonstrates its ability to precisely locate objects of interest and generate accurate descriptions through visual-language co-referring. We provide more results in Appendix \ref{app: quant}.

\begin{table}[t]
    \centering
    % \normalsize
    \fontsize{9pt}{\baselineskip}\selectfont
    \setlength{\tabcolsep}{2.8pt}
    \begin{tabular}{lcccc}
        \toprule
        Type  & $mAP$ & $AP_{50}$ & $AP_{75}$ & Mem.  \\
        \midrule 
        Resampler & 9.6 & 18.4 & 8.8 & 416G\\
        Down-sample projector (ours) & \textbf{28.4} & \textbf{43.1 }& \textbf{29.6} & 461M \\
        \bottomrule
    \end{tabular}
    \caption{Ablations on the different projectors with down-sampling. Mem. donates the memory consumption of this block including the parameters and forward/backward pass.}
    \label{aba: down}
\end{table}

\section{Conclusion}
In this study, we present Griffon v2, an innovative high-resolution multimodal model supporting resolutions up to 1K and facilitating visual-language co-referring. Our designed high-resolution structure directly extracts visual features and projects them into visual tokens with compression effectively and efficiently without division. Subsequently, we introduce a visual-language co-referring paradigm that accommodates locally cropped images, texts, and coordinates as prompts, offers diverse interactive capabilities, and mitigates the limitations of singular visual and textual prompting. Trained through our 3-stage end-to-end pipeline and the 12M multi-tasks and 900K instruction dataset, Griffon v2 surpasses expert models in object detection and counting tasks within a unified LVLM and demonstrates competitive performance across REC, REG, and phrase grounding tasks. Griffon v2 establishes a robust foundation for further exploration in the realm of intelligent multimodal systems with fine-grained perception and localization capabilities. We hope that the performance of Griffon v2 will instill confidence in the advancement of LVLMs.

\section{Acknowledgement}
This work was supported by Beijing Natural Science Foundation (L247028) and was in part by the Beijing Municipal Science and Technology Project (Z231100007423004), and National Natural Science Foundation of China (No.62276260, No.62472423).
{
    \small
    \bibliographystyle{ieeenat_fullname}
    \bibliography{main}
}

\clearpage
\setcounter{page}{1}
\maketitlesupplementary
\setcounter{section}{0}
This supplementary document extends our main paper by providing more details about the dataset we have constructed, the unified representation, the implementation, the Phase Grounding results, the REG results, the result analysis on REC, and more qualitative analyses that are not included in the main paper due to the length limit. \textbf{We will release the code and data upon publication of the paper.}

\section{Dataset Details}
\label{data details}
As demonstrated in the Training Pipeline of the main paper, we have collected and processed 12M data to build our pre-training and instruction-following dataset, with visual-language co-referring. In the section, We detail the data construction and the main data processing as below. 

\subsection{Pre-training Data Construction} 
\begin{figure}
    \centering
    \includegraphics[width=\linewidth]{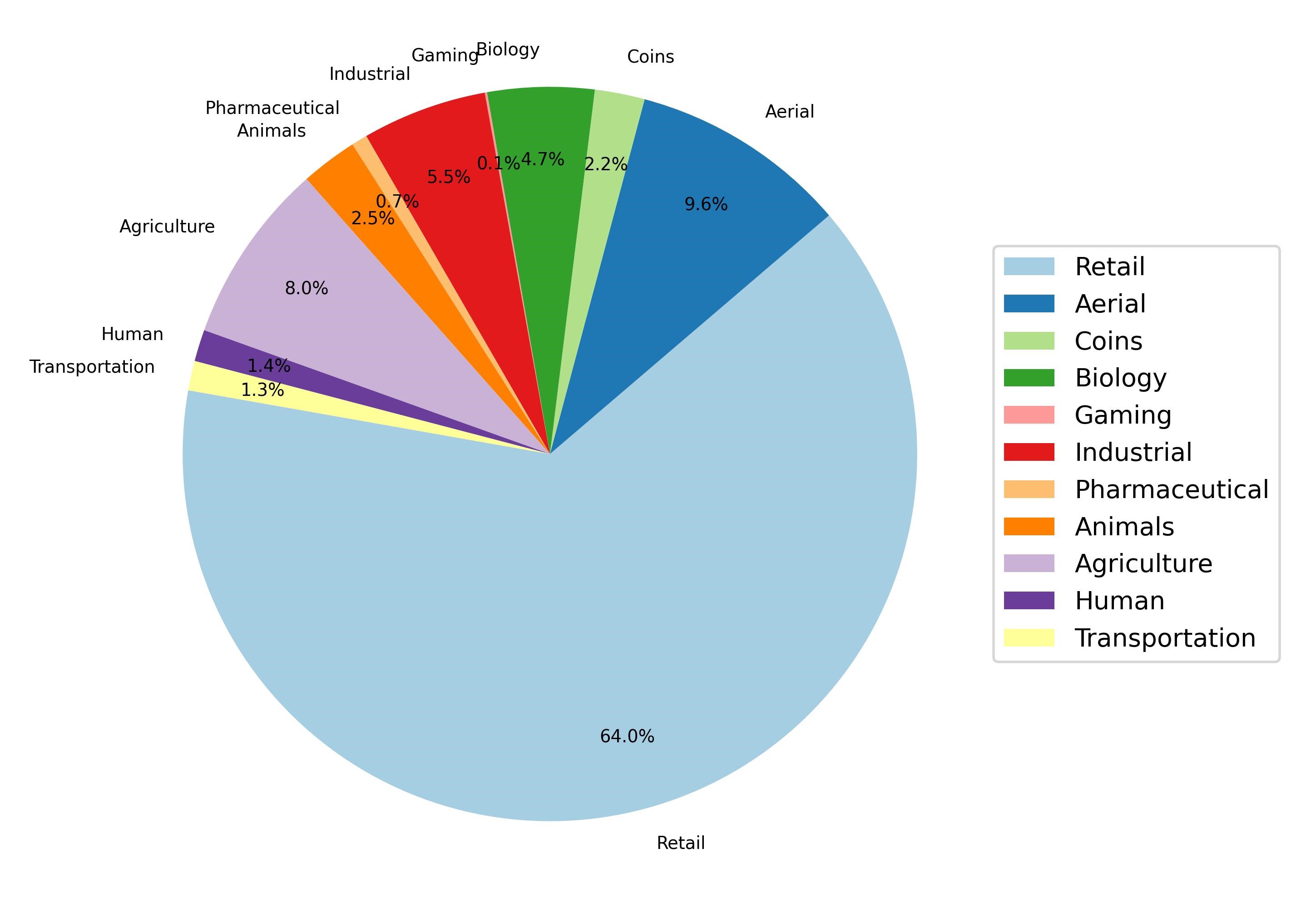}
    \caption{Data distribution of the self-collected counting data.}
    \label{fig: pie}
\end{figure}

\begin{figure}
    \centering
    \includegraphics[width=\linewidth]{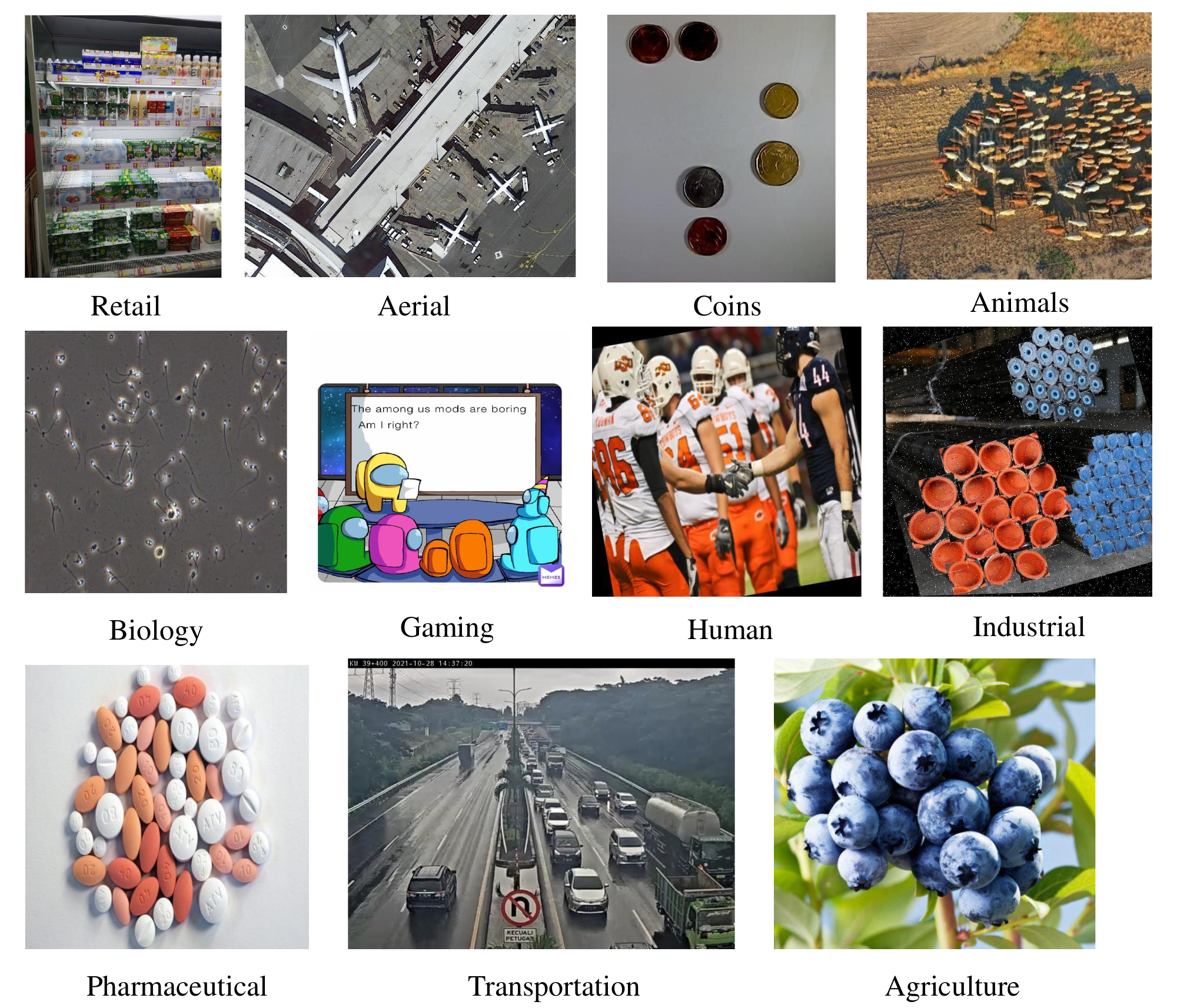}
    \caption{Data samples of the self-collected counting data.}
    \label{fig:self}
\end{figure}

To imbue the model with fine-grained perception and localization capabilities, and proficiency in visual-language co-referring, we curate a dataset of nearly 12 million localization-related instances with textual or visual reference. As illustrated in Table \ref{tab:pretraining data composition}, we encompass six localization-related tasks, transforming their respective datasets into a conversational style using task-specific prompts. The data from the object counting task are utilized for visual reference, while the remaining datasets serve for textual reference. Alongside the utilization of publicly available datasets, we have derived a counting subset comprising 416K instances from OpenImages v4 and a self-collected counting dataset comprising 266K instances. The counting subset filters out images lacking categories with instance numbers exceeding 5. The self-collected counting dataset integrates data from 11 domains, as depicted in Figure \ref{fig: pie} and Figure \ref{fig:self}. This broad domain coverage ensures the generalization of our model without succumbing to overfitting in any particular scenario.
\begin{table}[h]
    \centering
    \begin{tabular}{lll}
        \toprule
         Type & Dataset Name & Vol.\\
         \midrule
         \multirow{2}{*}{REC/REG} & Visual Genome & 3.6M \\
         & RefCOCO/+/g & 288K \\
         \midrule
         \multirow{2}{*}{Object Detection} & MSCOCO & 118K \\
         & Objects365 & 1.7M \\
         \midrule
         \multirow{3}{*}{Visual/Phrase Grounding} & LVIS & 361K \\
         & V3Det & 638K \\
         & Flickrs30K Entities & 427K \\
         \midrule
         \multirow{3}{*}{Object Counting} & CA-44 & 22K \\
         & OpenImages v4 & 416K \\
         & Self-collected & 266K \\
         \midrule
         Non-existing Judging & LVIS & 96K \\
         \bottomrule
    \end{tabular}
    \caption{The statistic of the composition of pre-training data.}
    \label{tab:pretraining data composition}
\end{table}

\subsection{Instruction-following Data Construction} In contrast to the extensive data providing wide knowledge used in the pre-training phase, we leverage a smaller subset of the multi-task localization pre-training data with a greater diversity of instruction prompts, exampled in Table \ref{tab:templates}, to enhance the model's understanding of intents. Instead of manually selecting subsets from various domains, we have opted for random sampling for both the visual grounding task and object counting task. We utilize the RefCOCO series for the REG/REC and MSCOCO for object detection. The data of each task realize a relative balance in terms of quantity. %The seeds and codes will be made available. %The overall proportion of the final instruction-following dataset, encompassing object detection, REC, visual grounding, object counting, and REG, follows a balanced sampling ratio inversely proportional to the data volume. The seeds and codes will be made available.

\begin{table*}[t]
    \centering
    \footnotesize
    \begin{tabular}{l|l}
        \toprule
        Task & Example prompts chosen from the instruction set\\
        \cmidrule(r){1-1}\cmidrule(r){2-2}
        \multirow{3}{*}{REC} & Where is $<$expr$>$ $<$image$>$? answer in [x0,y0,x1,y1] format. \\
        & I am looking for the position of $<$expr$>$ in $<$image$>$. Can you provide its coordinates? \\
        & Help me locate and determine the coordinates of $<$expr$>$ in $<$image$>$. \\
        \cmidrule(r){1-1}\cmidrule(r){2-2}
        \multirow{3}{*}{REG} & Please generate a distinguishing description for the region $<$region$>$ in the image $<$image$>$. \\
        & Describe the area $<$region$>$ in a unique way, given the picture $<$image$>$. \\
        & Create a one-of-a-kind description for the region $<$region$>$ found in the picture $<$image$>$. \\
        \cmidrule(r){1-1}\cmidrule(r){2-2}
        \multirow{6}{*}{\makecell[l]{Object\\Detection}} & \makecell[l]{Identify and locate all the objects from the category set in the image$<$image$>$. Please provide the coordinates for each \\detected object. The category set includes $<$category set$>$.}\\
        \cmidrule(r){2-2}
        & \makecell[l]{Examine the image$<$image$>$ for any objects from the category set. Report the coordinates of each detected object. The \\category set includes $<$category set$>$.}\\
        \cmidrule(r){2-2}
        & \makecell[l]{Locate and identify the objects from the category set in the image$<$image$>$. Output the coordinates of each detected object. \\The category set includes $<$category set$>$.}\\
        \cmidrule(r){1-1}\cmidrule(r){2-2}
        \multirow{3}{*}{\makecell[l]{Visual\\Grounding}} & Would you kindly provide the coordinates of $<$expr$>$ located in the picture $<$image$>$? \\
        & Find $<$expr$>$ in $<$image$>$ and share its coordinates with me.\\
        & In the given $<$image$>$, can you find $<$expr$>$ and tell me its coordinates?\\
        \cmidrule(r){1-1}\cmidrule(r){2-2}
        \multirow{6}{*}{\makecell[l]{Object\\Counting}} & Detect and record the positions of objects that bear resemblance to $<$region$>$ in this image.\\
        \cmidrule(r){2-2}
        & \makecell[l]{I want you to find all objects in the image$<$image$>$ that closely match the characteristics of $<$region$>$ and give me their \\coordinates.} \\
        \cmidrule(r){2-2}
        & \makecell[l]{Can you identify any objects that look like $<$region$>$ in this image$<$image$>$? Output their coordinates for closer inspection,\\ analysis, and comparison.} \\
        \bottomrule
    \end{tabular}
    \caption{Examples of task templates on different types of training data. The placeholders are explained as follows: ``$<$image$>$'' represents the input image, ``$<$expr$>$'' represents the expression describing the object, ``$<$category set$>$'' represents the categories to be detected, and ``$<$region$>$'' represents the textual coordinates of the region to be asked or the locally cropped image.}
    \label{tab:templates}
\end{table*}

\subsection{Data Processing} As previously mentioned, we consolidate six tasks into a unified instruction-answer format. For REC, REG, and object detection, we adopt the processing methodology introduced by Griffon-13B \cite{zhan2023griffon}, wherein raw annotations are directly transformed using randomly sampled instruction prompts. Regarding the visual grounding, object counting, and non-existing judging tasks, we initially convert detection-type annotations such as V3Det into instances, formulating one question for each category and enumerating all annotated categories for each image. Notably, in the case of non-existing judging data, we leverage the ``neg\_category\_ids'' annotated for each image, indicating categories unequivocally absent in that image. Subsequently, these data are integrated with randomly selected instruction templates.

\section{Instruction Template Examples}

In order to augment users' intent comprehension, we employ a diverse training instruction set along with a random sampling strategy. Here, we present a selection of task prompts utilized by Griffon v2 in Table \ref{tab:templates}. Each task encompasses hundreds of prompts generated by GPT-4 with specific requirements and illustrative examples. It is important to emphasize that Griffon v2 does not impose restrictions on users, allowing them the flexibility to employ their preferred natural language expressions.

\section{Unified Representation with VL Co-referring} 
\label{unified}
\begin{figure*}[t]
  \centering
  \includegraphics[width=\linewidth]{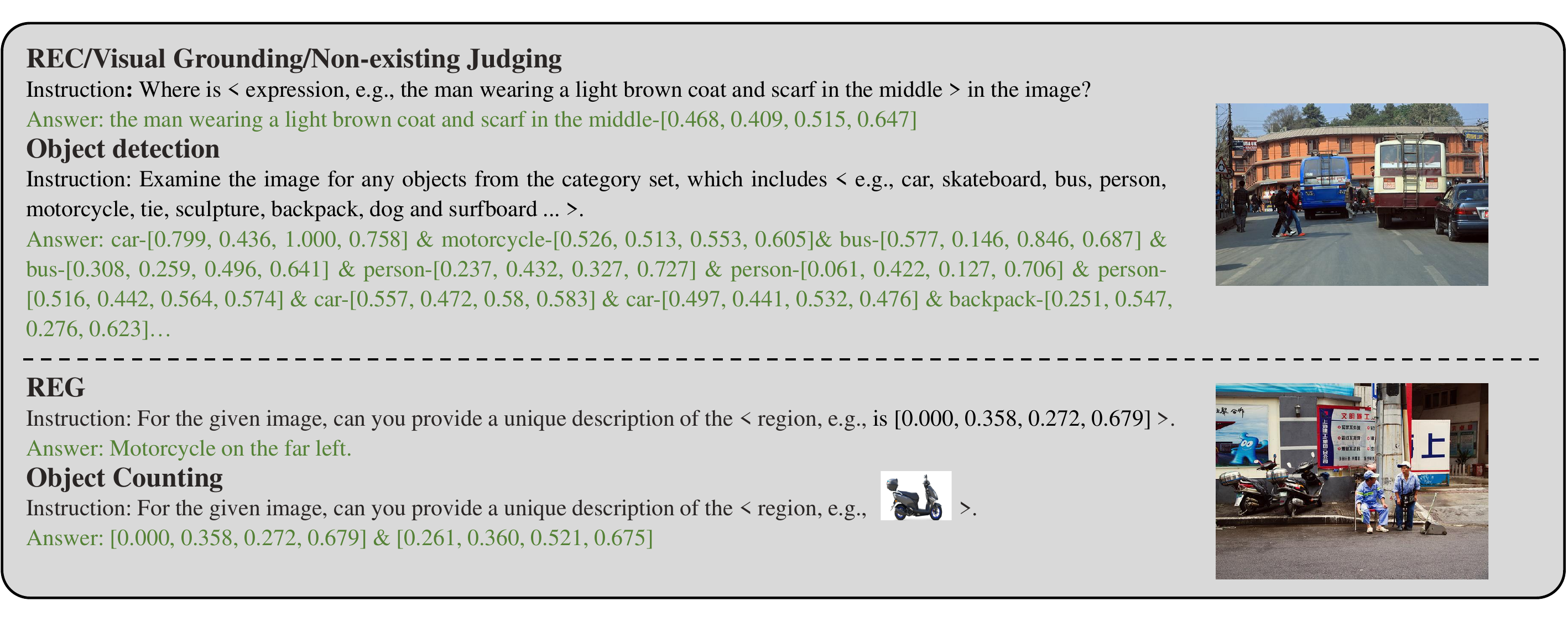}
   \caption{Examples of unified representation for each task. Object counting task utilizes visual referring without category information and directly outputs the object coordinates and corresponding number.}
   \label{fig: unifed}
\end{figure*}

Griffon v2 employs an enhanced unified input/output representation, building upon the framework introduced in Griffon-13B, in which the input is task-specific instruction and each instance in the output is formulated as ``expression-[x1, y2, x2, y2]''. The representation, as preliminarily illustrated in Figure \ref{fig: structure} from a referring perspective, has been upgraded to accommodate REG and object counting in addition to the previously supported REC, object detection, and visual/phrase grounding tasks. In Figure \ref{fig: unifed}, for the REG task, we refer to the question region with normalized 3-precision coordinates, ``[x1, y1, x2, y2]'' uniformly, seamlessly integrating it into the instruction, with the answer describing the region. Regarding object counting with visual referring, we initially employ the placeholder ``$<$region$>$'' in the instruction to denote the target. During training, this region is randomly selected from the bounding box annotation set of a specific category within the image, subsequently cropped out, and represented by the extracted token. The output sequence is the coordinates of detected instances concatenated with ``\&''. During inference, it's specified by the user with screenshots or target images. 

\section{Implementation Details}
\label{sup: imple}
% Model parameters and training hyperparameters are two key implementation details. In addition to the key settings we introduction in \ref{imple}, we list the full parameters in \ref{supp:hyper1} and \ref{supp:hyper2}. The training hyperparameters mainly follow the setting of LLaVA, and the max length increases to 4096 for higher resolution and longer texts. The epoch of stage 3 can increase to 3 following the full schedule for better performance.
Model parameters and training hyperparameters constitute crucial aspects of the implementation. Beyond the fundamental settings introduced in the main paper, the comprehensive lists of parameters are provided in Table \ref{supp:hyper1} and Table \ref{supp:hyper2}. The training hyperparameters predominantly adhere to the LLaVA configuration, with the maximum length extended to 4096 to accommodate higher-resolution images and longer texts. The total training time is about 40 NVIDIA A800 days similar to 100 patch-based Monkey \cite{li2023monkey} with more data.

\begin{table}[]
    \centering
    \begin{tabular}{clc}
        \toprule
        Layer & Parameters & Value \\
        \midrule
        \multirow{5}{*}{{Convolution}} & Stride & 2\\
        & Kernel & 3\\
        & Padding & 1\\
        & inchannel & 1024\\
        & outchannel & 5120\\
        \midrule
        \multirow{2}{*}{{Linear}}& inchannel & 5120 \\
        & outchannel & 5120 \\
        \bottomrule
    \end{tabular}
    \caption{Hyperparameters of the designed Down-sampling Projector.}
    \label{supp:hyper1}
\end{table}

\begin{table}[]
\begin{tabular}{c|ccc}
        \toprule
        Parameter & Stage-1 & Stage-2 & Stage-3\\
        \midrule
        batch size & 256 & 128 & 128\\
        lr & 1e-3 & 2e-5& 2e-5\\
        lr schedule & \multicolumn{3}{c}{cosine decay}\\
        lr warmup ratio & \multicolumn{3}{c}{0.03}\\
        weight decay & \multicolumn{3}{c}{0} \\
        epoch & \multicolumn{3}{c}{1}\\
        optimizer &\multicolumn{3}{c}{AdamW}\\
        DeepSpeed stage & \multicolumn{3}{c}{2}\\
        Max Length & 2048 & 4096 & 4096 \\
        \bottomrule
    \end{tabular}
    \caption{Hyperparameters of the training paradigm.}
    \label{supp:hyper2}
\end{table}

\section{Phase Grounding and REG results}
\label{sup:PG and REG}
\textbf{Phrase Grounding.} Phrase grounding task presents a greater challenge compared to the REC task and is evaluated on Flickr30K Entities \cite{plummer2015flickr30k}. Two evaluation protocols \cite{kamath2021mdetr} are employed, including the ANY-BOX protocol and MERGE-BOXES protocol. The ANY-BOX protocol focuses on the atomicity of each instance, while the MERGE-BOXES protocol evaluates whether the model identifies all referred objects with a merged box. Existing LVLMs are typically limited to the single referent scenario, tending to predict only one box per phase, thereby employing the MERGED-BOXES protocol. As shown in Table \ref{sup:pg}, Griffon v2 achieves state-of-the-art results in the ANY-BOX protocol and surpasses most specialists and generalists in the MERGE-BOX protocol, with more fine-grained boxes.
\begin{table}[t]
    \centering
    \fontsize{9pt}{\baselineskip}\selectfont
    \begin{tabular}{cccc}
        \toprule
        {Type} & {Model} & ANY & MERGED\\
        \midrule
         \multirow{3}{*}{\rotatebox{90}{Spec.}}
         & DDPN  & -  & 73.5 \\
         & VisualBert    &  71.3    & -\\
         & MDETR  &   83.4  & 83.8\\
         \midrule
         \multirow{5}{*}{\rotatebox{90}{Gen.}} & UniTAB  & - &  79.6\\
         & Ferret-13B & -  & \textbf{84.8}\\
         & Shikra-13B & -  & 78.4 \\
         & Griffon-13B &{84.2}  & 82.8\\
         & Griffon v2 & \textbf{84.8} & 83.1\\
         \bottomrule
    \end{tabular}
    \caption{Phrase grounding results on Flickr30K Entities\cite{plummer2015flickr30k} test set. Spec. represents specialists, while Gen. represents generalists.}
    \label{sup:pg}
\end{table}

\textbf{REG.} REG aims to generate concise descriptions for specified objects based on their region locations. We input textual coordinates for object referring and tests on the RefCOCOg \cite{nagaraja2016modeling} val set. As illustrated in Table \ref{main:reg}, in contrast to KOSMOS-2 \cite{peng2023kosmos}, which uses learnable embeddings for referring, we achieve superior performance in CIDEr, concentrating on the semantic similarity, while the Meteor focuses more on the accuracy of wording, making it less suitable for the open-ended description generation of LLMs.

\begin{table}[t]
    \centering
    \fontsize{9pt}{\baselineskip}\selectfont
    \begin{tabular}{clcc}
        \toprule
        Type & Model & CIDEr & Meteor \\
        \midrule
        \multirow{3}{*}{\rotatebox{90}{Spec.}} &SLR\cite{yu2017slr} &66.2 & \textbf{15.9} \\
        &ASM\cite{wang2023asm} & 41.9 & 13.6 \\
        &Grit\cite{wu2022grit} & 71.6 & 15.2 \\
        \midrule
        & KOSMOS-2\cite{peng2023kosmos}& 60.3 & 12.2\\
        \multirow{-2}{*}{\rotatebox{90}{Gen.}}& Griffon v2 & \textbf{72.5} & 12.1\\
        \bottomrule
    \end{tabular}
    \caption{REG resuls on RefCOCOg \cite{nagaraja2016modeling}. }
    \label{main:reg}
\end{table}

\section{More Qualitative Results}
\label{app: quant}
\begin{figure*}
  \centering
  \includegraphics[width=\linewidth]{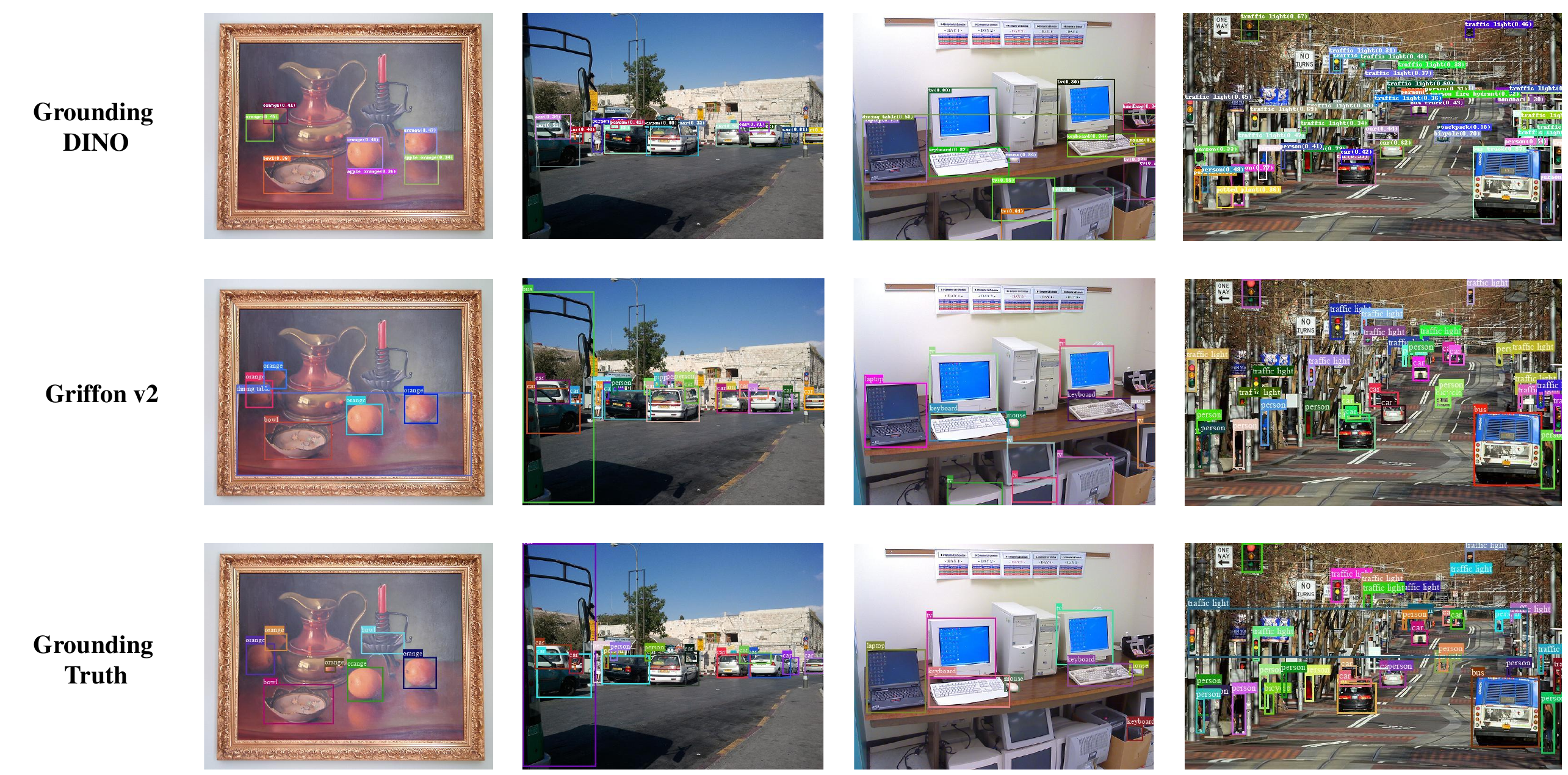}
   \caption{Comparison with Grounding DINO in object detection. Griffon v2 demonstrates a reduced occurrence of both missed detections (col. 2) and false positives (col. 1,3).}
   \label{fig: 1}
\end{figure*}

\begin{figure*}
  \centering
  \includegraphics[width=\linewidth]{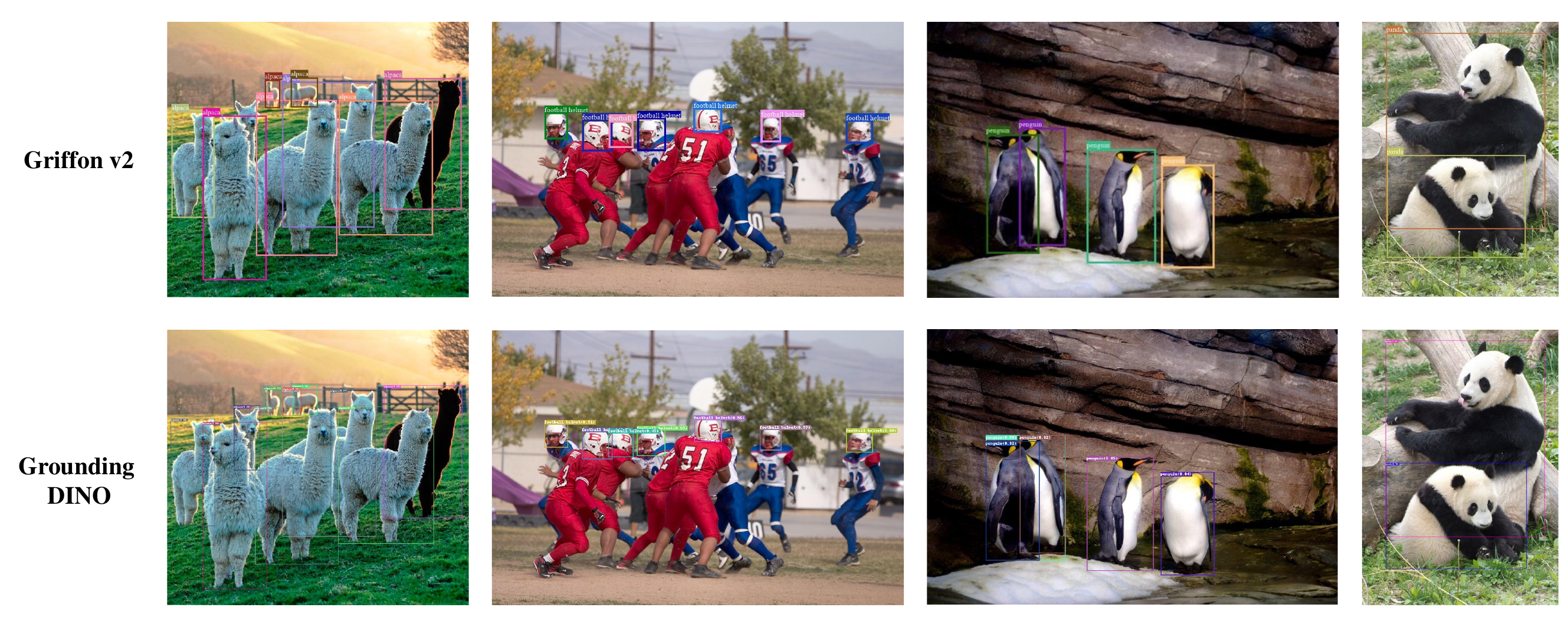}
   \caption{Comparison with Grounding DINO in visual grounding. Griffon v2 and Grounding DINO exhibit comparable visual grounding capabilities.}
   \label{fig: 2}
\end{figure*}

\begin{figure*}
  \centering
  \includegraphics[width=0.5\linewidth]{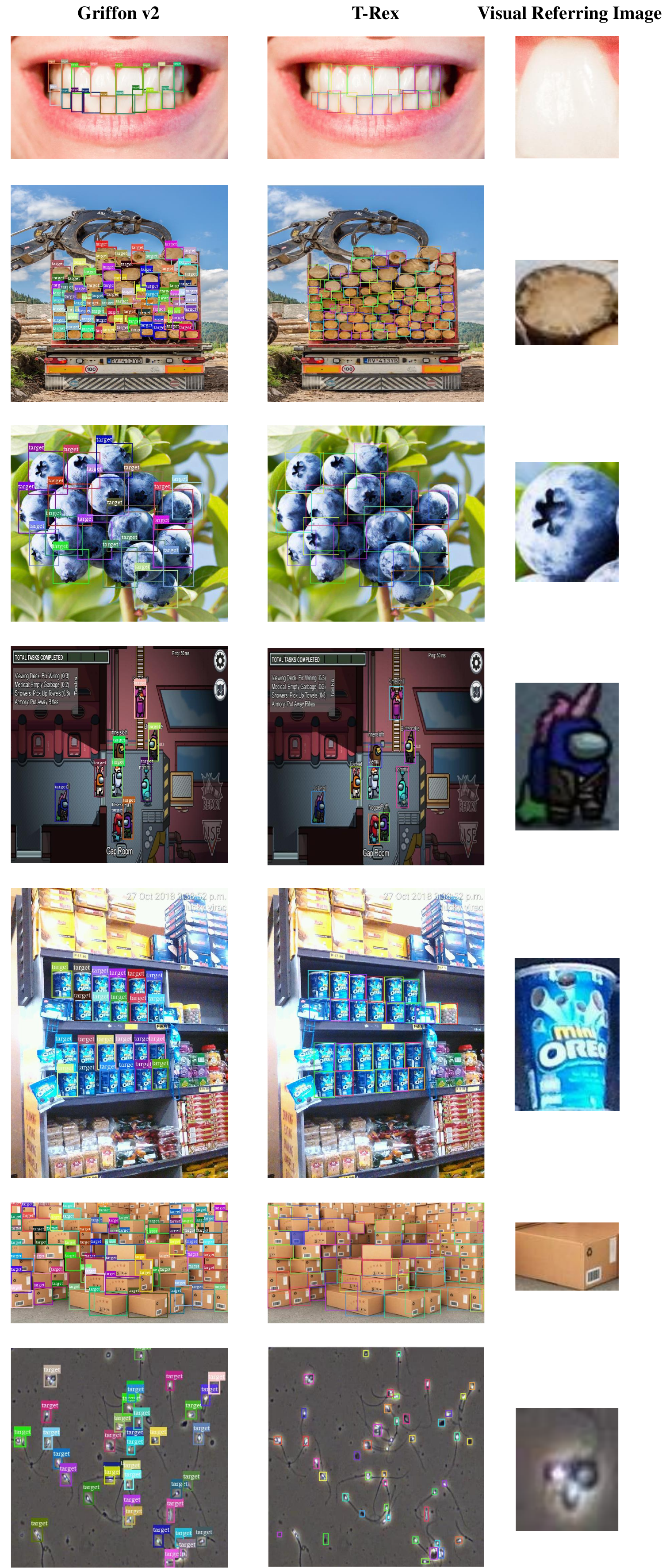}
   \caption{Comparison with T-Rex in object counting. Griffon v2 achieves counting proficiency with visual reference comparable to that of the expert model T-Rex.}
   \label{fig: 3}
\end{figure*}

To further demonstrate the performance of our Griffon v2, we provide more visualization results on object detection (Figure \ref{fig: 1}), visual grounding (Figure \ref{fig: 2}), and object counting (Figure \ref{fig: 3}), compared to some expert models.
\end{document}